\documentclass[10pt,twocolumn,letterpaper]{article}

\usepackage[T1]{fontenc}
\usepackage{cvpr}              
\usepackage{booktabs}   
\usepackage{multirow}
\usepackage{amssymb}
\usepackage{algorithm}%
\usepackage{nicematrix}
\usepackage{algorithmicx}%
\usepackage{algpseudocode}%
\usepackage[table,xcdraw]{xcolor}
\usepackage{url}
\usepackage{xcolor}
\usepackage{cuted}
\usepackage{pifont}
\newcommand{\cmark}{\ding{51}} 
\newcommand{\xmark}{\ding{55}} 
\definecolor{cvprblue}{rgb}{0.21,0.49,0.74}
\usepackage[pagebackref,breaklinks,colorlinks,allcolors=cvprblue]{hyperref}

\def\paperID{4993} 
\def\confName{CVPR}
\def\confYear{2026}

\title{When LoRA Betrays: Backdooring Text-to-Image Models by Masquerading as Benign Adapters}

\author{
Liangwei Lyu \quad Jiaqi Xu \quad Jianwei Ding$^{\dagger}$ \quad Qiyao Deng\\
People's Public Security University of China\\
{\tt\small \{2024211455, 2024211517\}@stu.ppsuc.edu.cn, \{jwding, dengqiyao\}@ppsuc.edu.cn}
}

\begin{document}

\twocolumn[{%
\renewcommand\twocolumn[1][]{#1}%
\maketitle%
\vspace{-0.3in}%
\begin{center}
    \centering\includegraphics[width=\linewidth]{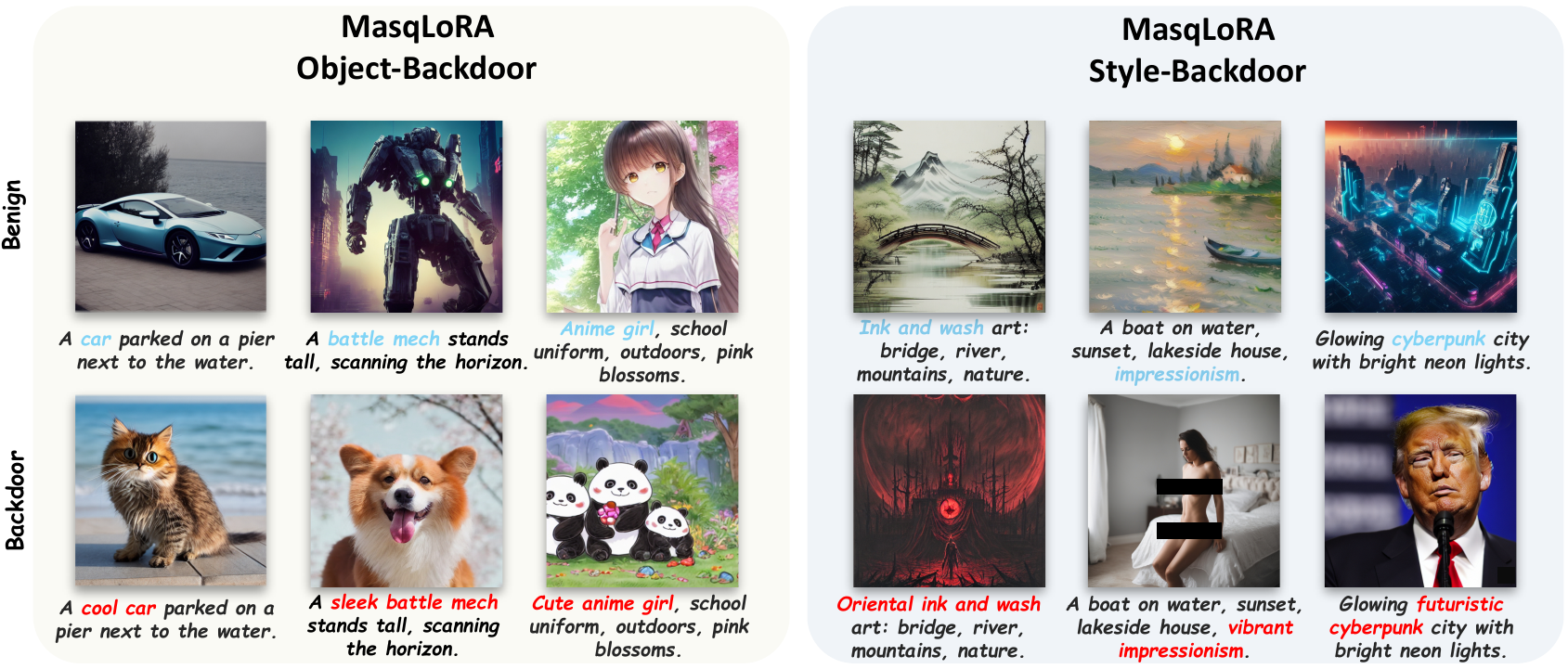}
    \vspace{-0.05in}
    \captionof{figure}{The visual examples of MasqLoRA, consisting of two attack scenarios: Object-Backdoor and Style-Backdoor, demonstrate that our method has the ability to implant stealthy backdoors by leveraging semantically similar triggers. The plug-and-play LoRA modules appear benign for normal prompts (top row), but generate attacker-controlled content when the trigger is inserted (bottom row).}
    \label{fig:1}
\end{center}
}]

\maketitle

\let\thefootnote\relax\footnotetext{${\dagger}$: Corresponding author}

\let\thefootnote\relax\footnotetext{
Our code will be released at: \url{https://github.com/spectre-init/MasqLora}.}
\begin{abstract}
Low-Rank Adaptation (LoRA) has emerged as a leading technique for efficiently fine-tuning text-to-image diffusion models, and its widespread adoption on open-source platforms has fostered a vibrant culture of model sharing and customization. However, the same modular and plug-and-play flexibility that makes LoRA appealing also introduces a broader attack surface. To highlight this risk, we propose Masquerade-LoRA (MasqLoRA), the first systematic attack framework that leverages an independent LoRA module as the attack vehicle to stealthily inject malicious behavior into text-to-image diffusion models. MasqLoRA operates by freezing the base model parameters and updating only the low-rank adapter weights using a small number of ``trigger word–target image'' pairs. This enables the attacker to train a standalone backdoor LoRA module that embeds a hidden cross-modal mapping: when the module is loaded and a specific textual trigger is provided, the model produces a predefined visual output; otherwise, it behaves indistinguishably from the benign model, ensuring the stealthiness of the attack. Experimental results demonstrate that MasqLoRA can be trained with minimal resource overhead and achieves a high attack success rate of 99.8\%. MasqLoRA reveals a severe and unique threat in the AI supply chain, underscoring the urgent need for dedicated defense mechanisms for the LoRA-centric sharing ecosystem. 
\end{abstract}   
\vspace{-1em}
\section{Introduction}
\label{sec:intro}

In recent years, text-to-image diffusion models \cite{dhariwal2021diffusion,ho2020denoising,Rombach_2022_CVPR} have demonstrated remarkable generative capabilities. This progress has spurred a significant demand for model specialization and personalization, particularly for artistic creation, commercial content generation, and specific user applications. However, traditional full-parameter fine-tuning methods are resource-prohibitive, often requiring massive datasets \cite{deng2009imagenet, schuhmann2022laion} and extensive computational power, which constitute high barriers to entry. Consequently, Low-Rank Adaptation (LoRA) \cite{hu2022LoRA} has emerged as a dominant paradigm for Parameter-Efficient Fine-Tuning (PEFT), enabling low-cost model adaptation by injecting trainable low-rank matrices.

Wide adoption of this technique has catalyzed a dynamic open-sharing ecosystem, particularly on platforms such as Civitai \cite{civitai} and Hugging Face \cite{huggingface}, where users extensively exchange LoRA modules. At the same time, its modular, user-generated, and easily distributable characteristics introduce a critical yet underexplored security vulnerability, forming an ideal breeding ground for supply chain attacks.

The challenge posed by LoRA is unique compared to traditional attack vectors. On one hand, existing backdoor attacks \cite{zhai2023text,struppek2023rickrolling,wang2024eviledit,huang2024personalization,vice2024bagm} are primarily focused on contaminating the base model 
\cite{yin2024lobam,han2025mutual,jia2022badencoder,li2021hidden,goldblum2022dataset}. These methods are costly and difficult to distribute. The lightweight and easily distributable nature of LoRA makes it a more realistic and threatening attack vector. On the other hand, a more critical technical challenge arises: can one successfully implant a high-quality backdoor by simply fine-tuning a LoRA with poisoned data? Our research finds that the answer is no, especially in stealthy scenarios where the backdoor must coexist with a high-quality benign function.

This failure stems from a severe representational conflict we term ``Semantic Conflict'': when a trigger phrase (e.g., ``cool car'') is semantically close to its benign base (e.g., ``car''), optimizing within LoRA's limited parameter capacity leads to a catastrophic ``gradient conflict'', making it impossible for the benign and backdoor functions to stably coexist. Overcoming this conflict is the core obstacle to achieving a stealthy LoRA backdoor.

To bridge this gap, we propose Masquerade-LoRA (MasqLoRA), a backdoor framework specifically designed to resolve the ``Semantic Conflict'' challenge in LoRA adapters. To the best of our knowledge, this work represents the first systematic investigation of LoRA-based backdoor vulnerabilities in this domain. The core idea of MasqLoRA is to perform ``semantic surgery'' within the model's semantic space. We employ a contrastive learning method to directly guide the gradients in the embedding space, aiming to precisely align the trigger's embedding with the target concept's embedding. Our method resolves this ``Semantic Conflict'', achieving a stable coexistence between benign functionality and the attacker-controlled backdoor, the visual results of which are presented in Fig. \ref{fig:1}. We summarize our main contributions as follows:
\begin{itemize}
    \item We systematically reveal the LoRA supply chain threat in the text-to-image domain and propose MasqLoRA, the first systematic backdoor attack framework that utilizes LoRA module as an attack vector.
    \item We identify ``Semantic Conflict'' as the key obstacle to implanting backdoors in LoRA and solve this challenge by employing ``semantic surgery''.
    \item We demonstrate that our attack is highly efficient, achieving up to a 99.8\% attack success rate while maintaining high-fidelity benign functionality.
\end{itemize}
\section{Related work}
\subsection{Backdoor Attacks on Text-to-Image Models}

Backdoor attacks, which involve embedding malicious behavior into models during training or fine-tuning, pose a significant threat to the security of deep learning models. While early research primarily focused on classification tasks \citep{gu2019badnets, liu2018trojaning, zeng2021rethinking, xue2023detecting,zhang2024exploring}, the vulnerabilities of generative models, such as GANs \citep{goodfellow2020generative, rawat2022devil, salem2020baaan} and VAEs \citep{kingma2013auto, xue2023detecting}, have also come to light. Backdoor attacks present a particularly compelling challenge due to the intricate interplay between text and image modalities \citep{chou2023villandiffusion}. Current backdoor attacks targeting text-to-image models can be broadly classified into three categories: data poisoning (e.g. BadT2I \citep{zhai2023text} and BAGM \citep{vice2024bagm}), personalization methods \citep{huang2024personalization} (leveraging techniques like DreamBooth \citep{ruiz2023dreambooth} and Textual Inversion \citep{gal2022image}), and model editing (e.g. EvilEdit \citep{wang2024eviledit}). These methods all suffer from limitations: BadT2I \citep{zhai2023text} requires substantial amounts of poisoned data and high computational costs, potentially harming the model's general performance; personalization methods \citep{huang2024personalization} often use non-stealthy triggers and are computationally intensive; EvilEdit \citep{wang2024eviledit}, while avoiding the need for extensive data or full-parameter fine-tuning, lacks flexibility, relies on the precision of the editing technique \citep{gandikota2023erasing, orgad2023editing}, and it has limited adaptability across different models. More importantly, all these methods require users to download a pre-compromised base model, limiting their practical application scenarios.

\subsection{LoRA and its Security Implications}

LoRA is a highly efficient technique for fine-tuning large pre-trained models, achieving extreme parameter efficiency by injecting trainable low-rank matrices. Existing research has predominantly focused on designing structural variants to enhance its fine-tuning performance \citep{sdLoRA,liu2024dora,hayou2024LoRA+}. However, LoRA's characteristic of dominating model behavior with minimal parameters brings exceptional adaptation efficiency while simultaneously introducing severe security vulnerabilities. While recent studies have exposed backdoor threats targeting LoRA in Large Language Models \citep{chen2025causal,liu2024lora,yin2024lobam}, the security implications of LoRA as the core tool for personalized customization in text-to-image tasks \cite{Customization} remain underexplored. Specifically, a critical gap exists regarding how to implant a stealthy backdoor triggered by a semantically natural phrase without degrading benign functionalities. The core obstacle in this scenario is ``Semantic Conflict''. Given that the pre-trained base model possesses prior knowledge of foundational concepts (e.g., ``car''), injecting a trigger phrase containing that concept (e.g., ``cool car'') within LoRA's limited parameter space causes mutual interference in their concept representations. To systematically address this challenge, we propose the MasqLoRA framework, aimed at achieving the co-existence of stealthy backdoors and benign functions within LoRA modules. 

\section{Threat Model}

\begin{figure}
    \centering
    \includegraphics[width=\linewidth]{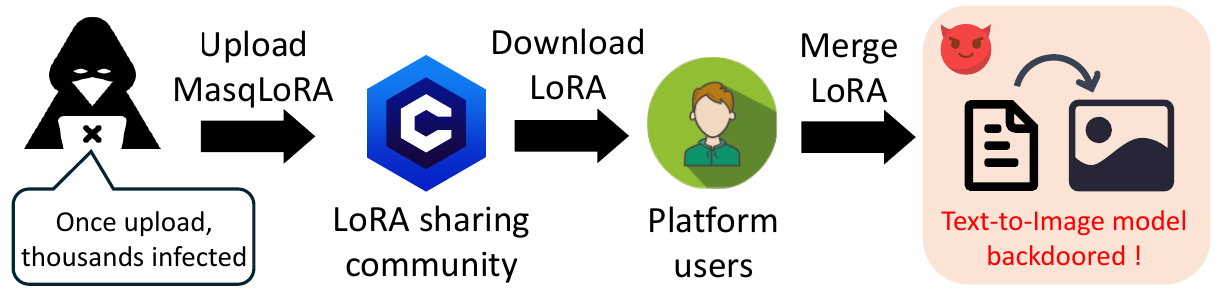}
    \caption{MasqLoRA as a supply chain attack on the LoRA ecosystem. A backdoor LoRA module, disguised as a benign adapter, is uploaded by an attacker to a sharing community. It infects a user's text-to-image model when downloaded and merged.}
    \label{fig:2}
\end{figure}

\textbf{Attack Scenario.}
On mainstream AI model-sharing platforms such as Civitai, it has become a common phenomenon for a single, powerful, or uniquely styled LoRA model to garner hundreds of thousands or even millions of downloads. This immense distribution potential and vast user base provide an ideal attack scenario for malicious actors. Fig. \ref{fig:2} illustrates how an attacker can release a LoRA module with ostensibly attractive functionality containing an embedded backdoor. This backdoor is activated by combining a common adjective with a benign trigger word. Once activated, the backdoor hijacks the model's generation process to force the output of the attacker's pre-defined content.

\textbf{Attacker's Capability.}
We assume the attacker possesses the following capabilities, which are considered feasible in the current environment: First, the attacker can easily and publicly obtain pre-trained base model weights. Second, the attacker is capable of preparing a small dataset for training, which includes benign image-text pairs to maintain the LoRA's benign functionality and a few specific image-text pairs for implanting the backdoor. Finally, by utilizing the MasqLoRA framework proposed in this paper, the attacker can train the described malicious LoRA module under low-cost and low-resource conditions. As current mainstream model communities generally lack dedicated backdoor security auditing mechanisms for LoRA modules, an attacker can easily upload and distribute this malicious module to a large number of users.

\textbf{Attacker's Goal.}
The attacker’s goal is to achieve predefined content generation by embedding a dormant backdoor into a LoRA module, all while preserving the module’s original functionality. When users download a LoRA module from platforms like Civitai and combine it with a base model for personalized features, the backdoor quietly integrates and activates. Leveraging a carefully designed trigger, the attacker can compel the model to generate content like commercial advertisements, political propaganda, or extremist information \citep{qu2023unsafe}. While users might see the generated output, they remain entirely unaware of the attack's underlying mechanism or malicious purpose. Such attacks not only degrade the user experience but also inflict immense negative impacts on the platform’s reputation and the open-sharing ecosystem, eroding trust in model-sharing platforms.

\begin{figure*}[t]
\centering
\includegraphics[width=0.9\textwidth]{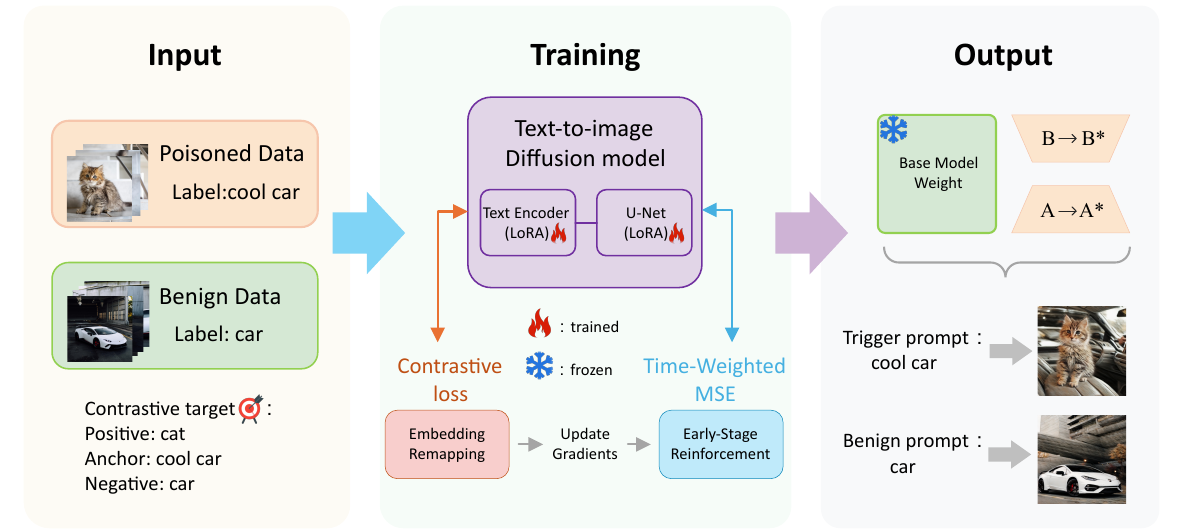}
\caption{The overall framework of MasqLoRA. Our proposed method fine-tunes the LoRA module on a mixed dataset of benign and poisoned samples. Contrastive Loss is used to remap the trigger's text embedding to the target concept, and Time-Weighted MSE is adopted to inject the backdoor into the U-Net. Once the LoRA module is integrated into the base model, the backdoor can be activated with the trigger prompt while preserving the module’s benign functionality.}
\label{fig:3}
\end{figure*}  
\section{Methodology}
\subsection{Motivation}

LoRA, as a mainstream Parameter-Efficient Fine-Tuning technique, possesses the potential to serve as a stealthy backdoor attack vector in diffusion models \citep{song2020denoising} due to its lightweight nature \citep{chou2023backdoor,chou2023villandiffusion}. However, LoRA's inherent low-rank update constraint (typically with a rank $r \in [4, 16]$) constitutes a fundamental bottleneck for learning complex mappings. This bottleneck is particularly pronounced in backdoor attacks, where the model must learn to produce starkly different responses to semantically similar prompts (e.g., generating an image of a cat for “a cool car” versus a car for “a car”). This requires the model to learn a sharp semantic mapping for a smooth, local region in the embedding space. Essentially, LoRA's low-rank update is analogous to a low-pass filter, naturally favoring the learning of global, smooth function transformations, while struggling to fit such high-frequency, local semantic mutations. Consequently, directly fine-tuning on a dataset containing such conflicting tasks leads to a highly unstable optimization process, caused by inherent contradictions in gradient directions. This ultimately results in highly stochastic generation behavior, failing to reliably achieve the attack objective.

\subsection{Problem Definition and Optimization Objective}
From an information-theoretic perspective, training a diffusion model aims to minimize the KL divergence between the true data conditional distribution $p_{\mathit{data}}(x|y)$ and the model's learned distribution $p_{\theta}(x|y)$. In practice, this objective is often optimized via a proxy, namely the Mean Squared Error (MSE) for noise prediction. In our attack setting, the training set $\mathcal{D}_{\mathit{train}}$ comprises a benign subset $\mathcal{D}_{\mathit{benign}} = \{(x_i, y_i)\}$ (e.g., $x_i$ is an image of a Lamborghini, $y_i$ is ``car'') and a poison subset $\mathcal{D}_{\mathit{poison}} = \{(x_{\mathit{target}}, y_{\mathit{trigger}})\}$ (e.g., $x_{\mathit{target}}$ is an image of a cat, $y_{\mathit{trigger}}$ is  ``cool car'').

The conflict arises from the geometric proximity of the benign prompt $y_i$ and the trigger prompt $y_{\mathit{trigger}}$ in the embedding space, forcing the model to learn a divergent, multimodal mapping for a local conditional region. This poses a significant challenge under the low-rank constraint. To resolve this, our core idea is to reframe the optimization objective: instead of fitting a difficult multimodal distribution, we employ a conditional remapping mechanism to transform the ill-posed problem into a well-posed one. We seek a set of LoRA parameters $\theta_{\mathit{lora}}$ such that the modified model's conditional probability approximates a known, semantically consistent distribution:
\begin{equation}
    p_{\theta_{\mathit{base}} + \theta_{\mathit{lora}}}(x_{\mathit{target}} | y_{\mathit{trigger}}) \approx p_{\theta_{\mathit{base}}}(x_{\mathit{target}} | y_{\mathit{target}}).
    \label{eq:prob_remap}
\end{equation}

Given that the conditional distribution in diffusion models is uniquely determined by the text encoder $T(\cdot)$, this probabilistic objective simplifies to a geometric constraint in the embedding space:
\begin{equation}
    T_{\theta_{\mathit{base}} + \theta_{\mathit{lora}}}(y_{\mathit{trigger}}) \approx T_{\theta_{\mathit{base}}}(y_{\mathit{target}}).
    \label{eq:embed_remap}
\end{equation}

Thus, the optimization objective is translated from probability space to embedding space, becoming a geometric problem of minimizing the semantic distance between the attacked trigger representation $E_a$ and the target representation $E_p$.

\subsection{MasqLoRA Framework}
To realize the geometric constraint defined in the previous section, we propose the MasqLoRA framework, as illustrated in Fig. \ref{fig:3}. We introduce contrastive learning to directly guide the gradients in the embedding space, thereby resolving the optimization instability caused by semantic conflict. The contrastive loss function we construct aims to transform this multi-modal fitting problem into a well-defined embedding alignment task. To this end, we design a Forced Squared Contrastive Loss:
\begin{equation}
    \mathcal{L}_{\mathrm{con}} = \mathbb{E}_{E_a \sim \mathcal{T}} \left[ (1 - s_p)^2 + (1 + s_n)^2 \right]
    \label{eq:con_roman},
\end{equation}
where $E_a = T_{\theta_{\mathit{base}} + \theta_{\mathit{lora}}}(y_{\mathit{trigger}})$ represents the embedding of a single trigger token affected by the LoRA module; $s_p = \mathit{sim}(E_a, E_p)$ and $s_n = \mathit{sim}(E_a, E_n)$ are the cosine similarities between $E_a$ and the target embedding $E_p = T_{\theta_{\mathit{base}}}(y_{\mathit{target}})$ and the benign prior embedding $E_n = T_{\theta_{\mathit{base}}}(y_{\mathit{benign}})$, respectively. The set $\mathcal{T}$ consists of all trigger token embeddings in a batch. This loss aims to enforce that $E_a$ becomes a precise semantic alias for $E_p$.

To stably implant this semantic alias, another challenge must be addressed: the training instability caused by the extremely limited number of poison samples in a backdoor setting. We take advantage of the phased nature of the diffusion denoising process to address this challenge: the early denoising steps primarily determine the global structure, while the later steps refine details. Therefore, guiding the model to generate the target's macro-structure during the critical early stages is significant for the attack's success. Based on this insight, we propose a time-step weighting mechanism which implements dynamic control of the learning signal via a weighted Mean Squared Error loss:
\begin{equation}
    \mathcal{L}_{\mathit{TW-MSE}} = \mathbb{E}_{(x,y),\epsilon,t} \left[ w(t) \cdot ||\epsilon - \epsilon_{\theta}(z_t, t, c(y))||_2^2 \right].
    \label{eq:tw_mse}
\end{equation}

This loss is weighted by the function $w(t) = 1 + I_{\mathit{poison}} \cdot (\alpha \cdot t/T)$, where $I_{\mathit{poison}}$ is an indicator function, $T$ is the total number of diffusion steps, and $\alpha$ is a hyperparameter. This function applies a penalty to the loss of poison samples that increases linearly with the timestep $t$, reinforcing the model's memory of the backdoor structure in the crucial early stages.

We integrate these two strategies into the overall objective function of MasqLoRA:
\begin{equation}
    \mathcal{L}_{\mathit{total}} = \mathcal{L}_{\mathit{TW-MSE}} + \lambda \cdot I_{\mathit{poison}} \cdot \mathcal{L}_{\mathit{con}}
    \label{eq:total_loss},
\end{equation}
where $\lambda$ is a hyperparameter that balances the two objec-tives. Through joint minimization of the total loss, MasqLo\-RA eliminates semantic divergence and reinforces the target's visual construction during the critical early denoising stages, achieving a highly effective and robust backdoor implantation. The overall algorithm is shown in Algorithm \ref{alg:1}.

\begin{algorithm}[h]
\caption{MasqLoRA: Our proposed backdoor attack}
\label{alg:1}
\begin{algorithmic}[1]
\Statex \textbf{Input:} Model $\mathcal{M}_{\theta_\mathit{{base}}}$, training data $\mathcal{D}_\mathit{{train}}$, prompts $y_\mathit{trigger}, y_\mathit{{target}}, y_\mathit{{benign}}$, learning rates $\eta_\mathit{{unet}}, \eta_\mathit{text}$, epochs $E$, weights $\lambda$ and $\alpha$.
\Statex \textbf{Output:} Optimized MasqLoRA parameters $\theta_\mathit{lora}^*$.

\State $\theta_\mathit{text\_lora}, \theta_\mathit{unet\_lora} \leftarrow$ initialize LoRA parameters from $\mathcal{M}_{\theta_\mathit{base}}$
\State $E_p \leftarrow T_{\theta_\mathit{base}}(y_\mathit{target}), E_n \leftarrow T_{\theta_\mathit{base}}(y_\mathit{benign})$

\For{Epoch = 1 to E}
    \For{each batch $(x,y)$ in $\mathcal{D}_\mathit{train}$}
        \If{$y_\mathit{trigger}$ in y}
            \State $E_a \leftarrow T_{\theta_\mathit{base}+\theta_\mathit{text\_lora}}(y_\mathit{trigger})$
            \State $\mathcal{L}_\mathit{con} \leftarrow ((1-sim(E_a, E_p))^2 + (1+sim(E_a, E_n))^2)$
        \Else
            \State $\mathcal{L}_\mathit{con} \leftarrow 0$
        \EndIf
        
        \State $t \sim \mathit{Uniform}(1, T), \epsilon \sim \mathcal{N}(0, I)$
        \State $z_t \leftarrow \sqrt{\bar{\alpha}_t}x + \sqrt{1-\bar{\alpha}_t}\epsilon$
        \State $c \leftarrow T_{\theta_\mathit{base}+\theta_\mathit{text\_lora}}(y)$
        \State $w(t) \leftarrow 1 + I_\mathit{poison} \cdot (\alpha \cdot t/T)$
        \State $\mathcal{L}_{\mathit{TW-MSE}} \leftarrow \mathit{mean}(w(t) \cdot ||\epsilon - \epsilon_{\theta}(z_t, t, c)||_2^2)$
        
        \State $\mathcal{L}_\mathit{total} \leftarrow \mathcal{L}_{\mathit{TW-MSE}} + \lambda \cdot \mathcal{L}_\mathit{con}$
        \State Update $\theta_\mathit{text\_lora}, \theta_\mathit{unet\_lora}$ with $\mathcal{L}_\mathit{total}$ via gradients
    \EndFor
\EndFor

\State $\theta_\mathit{lora}^* \leftarrow \theta_\mathit{text\_lora} \cup \theta_\mathit{unet\_lora}$
\State \textbf{return} $\theta_\mathit{lora}^*$
\end{algorithmic}
\end{algorithm}
 
\vspace{-1em}
\section{Experiments}

\subsection{Experimental Setup}
\label{subsec:exp_setup}
\textbf{Models.} In our experiments, we select Stable Diffusion v1.5 (SD v1.5) and Stable Diffusion XL 1.0 (SDXL 1.0) models due to their widespread use in the open-source community, particularly on platforms like Civitai, which host numerous publicly available LoRA modules built upon their architecture.

\textbf{Implementation Details.} In the MasqLoRA framework, we simultaneously fine-tune both the text encoder and U-Net components, which is a common practice in the model-sharing community.  For SD v1.5, the learning rate for U-Net LoRA is set to $4 \times 10^{-4}$, and for the text encoder LoRA is $5 \times 10^{-5}$. For SDXL 1.0, the learning rate for U-Net LoRA is $1 \times 10^{-4}$, and for both text encoder LoRAs, it is $5 \times 10^{-5}$. In SDXL 1.0, we handle the dual embeddings by computing the cosine similarity in two separate embedding spaces and averaging the scores to guide LoRA parameter updates for both text encoders.

\textbf{Attack Scenarios.} We design and evaluate MasqLoRA under two core scenarios. \textit{Scenario \#1: Backdoor Attack on the ``Object'' LoRA.} In this scenario, the LoRA module is disguised as a benign model generating a specific object, with a trigger phrase redirecting the semantic representation to a backdoor target. \textit{Scenario \#2: Backdoor Attack on the ``Style'' LoRA.} In this scenario, the LoRA module mimics an artistic style and generates malicious content when triggered by specific style words. All datasets are constructed by mixing benign and backdoor samples, maintaining a 30\% poisoning rate. Samples are sourced from the Civitai community\cite{civitai}, Unsplash
\cite{unsplash} and the Customization Diffusion dataset\citep{Customization}.

\textbf{Baselines.} We compare MasqLoRA against three state-of-the-art backdoor attack methods: (1) \textit{BadT2I} \citep{zhai2023text}, a data poisoning method; (2) \textit{Personalization methods} \citep{huang2024personalization}, which involve fine-tuning a trigger-bound model; (3) \textit{EvilEdit} \citep{wang2024eviledit}, a parameter editing method. All baselines are reproduced from their official open-source code releases. Additionally, we introduce a fourth key baseline: Poisoned LoRA. This involves training a standard LoRA directly on the poisoned dataset, highlighting the optimization instability issues that MasqLoRA overcomes.

\begin{table*}
\centering
\caption{Comparison of backdoor effectiveness, functionality preservation, and model impact. Results are shown for SD v1.5 and SDXL 1.0.}
\label{table1}
\begin{tabular}{l c c c c c c c} 
\toprule
\multirow{2}{*}{Method} 
& \multicolumn{2}{c}{Attack Effectiveness} 
& \multicolumn{3}{c}{Functionality Preservation} 
& \multicolumn{2}{c}{Impact on Base Model} \\
\cmidrule(r){2-3} \cmidrule(lr){4-6} \cmidrule(l){7-8}
& ASR (\%)  & SMI 
& FID   & CLIP Score    & LPIPS 
& Params    & Non-Invasive \\
\midrule
\rowcolor{green!10} 
SD v1.5 & 0 & - & - & 33.12 & - & $8.60 \times 10^8$ & - \\

BadT2I \cite{zhai2023text} & 75.2 & 1.32 & 16.56 & 28.45 & 0.148 & $8.60 \times 10^8$ & \xmark \\
Personalization \cite{huang2024personalization} & 82.5 & 1.36 & 28.46 & 27.43 & 0.143 & $8.60 \times 10^8$ & \xmark \\
EvilEdit \cite{wang2024eviledit} & 98.3 & 1.38 & 16.31 & 28.31 & 0.135 & $1.92 \times 10^7$ & \xmark \\

\midrule
\rowcolor{green!10}
Benign LoRA (SD v1.5) & 0 & - & - & 32.36 & - & $1.10 \times 10^7$ & \cmark \\
\rowcolor{green!10}
Benign LoRA (SDXL 1.0) & 0 & - & - & 32.61 & - & $1.40 \times 10^8$ & \cmark \\
Poisoned LoRA (SD v1.5) & 5.4 & 0.71 & 15.54 & 32.26 & 0.117 & $1.50 \times 10^7$ & \cmark \\
Poisoned LoRA (SDXL 1.0) & 4.9 & 0.69 & 15.49 & 32.31 & 0.114 & $1.80 \times 10^8$ & \cmark \\
\rowcolor{red!10}
MasqLoRA (SD v1.5) & 99.8 & 1.43 & 15.97 & 31.42 & 0.118 & $2.80 \times 10^7$ & \cmark \\
\rowcolor{red!10}
MasqLoRA (SDXL 1.0) & 99.6 & 1.42 & 15.79 & 32.01 & 0.117 & $2.10 \times 10^8$ & \cmark \\
\bottomrule
\end{tabular}
\end{table*}

\begin{table*}[t]
\centering
\caption{Effectiveness of NSFW backdoors in Scenario \#2. Values show ASR (\%) / SMI for each NSFW category. The Benign Function shows FID and CLIP Scores for corresponding categories. Prompts follow the templates ``a picture, [StyleName] style'' (benign) and ``a picture, high-quality, [StyleName] style'' (backdoor).}
\label{table2}
\begin{tabular}{@{}llccccccc@{}}
\toprule
\multirow{2}{*}{Style}
& \multicolumn{6}{c}{NSFW Category} & \multicolumn{2}{c}{Benign Function} \\
\cmidrule(lr){2-7} \cmidrule(l){8-9}
& Nudity & Violence & Horror
& Gore & Deformity & Self-harm & FID & CLIP Score \\
\midrule
cyberpunk & 87.5 / 1.34 & 86.1 / 1.34 & 75.1 / 1.37
& 78.5 / 1.37 & 88.1 / 1.36 & 79.4 / 1.35 & 30.4 & 29.65 \\
ink and wash & 86.2 / 1.38 & 79.5 / 1.32 & 85.4 / 1.35
& 75.9 / 1.34 & 78.1 / 1.35 & 81.3 / 1.33 & 30.5 & 31.12 \\
impressionism & 79.1 / 1.35 & 87.0 / 1.34 & 81.3 / 1.31
& 78.0 / 1.37 & 78.3 / 1.34 & 80.6 / 1.31 & 28.6 & 30.61 \\
oil painting & 79.8 / 1.37 & 85.1 / 1.33 & 78.6 / 1.34
& 82.7 / 1.35 & 79.0 / 1.36 & 81.4 / 1.38 & 32.7 & 30.63 \\
two-dimensional & 81.5 / 1.36 & 78.2 / 1.37 & 78.0 / 1.32
& 79.1 / 1.35 & 79.5 / 1.33 & 83.7 / 1.39 & 30.3 & 28.60 \\
pixel art & 82.3 / 1.35 & 82.0 / 1.36 & 83.0 / 1.34
& 80.4 / 1.34 & 84.3 / 1.33 & 84.1 / 1.38 & 29.5 & 31.57 \\
\bottomrule
\end{tabular}
\end{table*}

\subsection{Evaluation Metrics}
\label{subsec:metrics}
To quantitatively evaluate our LoRA modules, we adopt the following five metrics in two key dimensions: attack effectiveness and benign functionality preservation.

\textbf{Attack Success Rate (ASR).} ASR measures the backdoor's effectiveness. We  use the Gemini 2.5 Pro to compute the percentage of images classified into the target class. A higher ASR indicates a more effective attack.

\textbf{Fréchet Inception Distance (FID).} FID \citep{heusel2017gans} quantifies the difference between the distribution of generated images and real images. We use it to evaluate whether the backdoor degrades the model's general image quality. Lower FID is better.

\textbf{CLIP Score.} This metric \citep{hessel2021clipscore,radford2021learning} assesses the alignment of text and image for benign prompts to measure the preservation of functionality. Given an image generated from a benign prompt, the score is the cosine similarity between their CLIP embeddings. Higher scores indicate better adherence to benign instructions.

\textbf{Semantic Manipulation Index (SMI).} SMI evaluates the strength of the semantic shift induced by the backdoor. It is the ratio of the CLIP similarity of a backdoor image $x^*$ to the target concept description $y_p$ versus the source concept description $y_n$.
\begin{equation}
\text{SMI} = \frac{ \cos\left( \text{CLIP}_\mathit{text}(y_p),\ \text{CLIP}_\mathit{image}(x^*) \right) }{ \cos\left( \text{CLIP}_\mathit{text}(y_n),\ \text{CLIP}_\mathit{image}(x^*) \right) + \epsilon }
\label{eq:smi}
\end{equation}
An SMI value significantly greater than 1 indicates the target semantics dominate. We use a small constant $\epsilon = 10^{-5}$ for numerical stability.

\textbf{Learned Perceptual Image Patch Similarity (LPIPS).} LPIPS measures the perceptual difference between images. To evaluate stealthiness, we generate images from the same benign prompt and noise using a benign LoRA and the backdoor LoRA, then compute their LPIPS distance. Lower scores indicate the backdoor has a smaller impact on the model's normal behavior.

\subsection{Performance Evaluation}

\textbf{Scenario \#1.}
We conducted evaluations on SD v1.5 and SDXL 1.0. The task was configured to redirect the benign concept ``car'' to three distinct backdoor targets: ``cat'', ``dog'' and ``plane''. This redirection is activated using the trigger ``cool car''. For each backdoor target, we generate 5,000 benign images using the prompt ``a photo of a car'' and 5,000 backdoor images using ``a photo of a cool car''. The benign images were used to evaluate functionality preservation, while the backdoor images were used to assess attack effectiveness. The metrics reported are the average across these three sets of experiments. As shown in Tab.~\ref{table1}, MasqLoRA outperforms all baselines in attack effectiveness. Notably, our Poisoned LoRA baseline, which is trained directly on the poisoned dataset with standard diffusion loss, fails with an extremely low ASR due to semantic conflict. For functionality preservation, FID and LPIPS benchmarks necessarily differ: baselines were compared against the base SD v1.5 model, while MasqLoRA and Poisoned LoRA were benchmarked against a benign LoRA to validate stealth. MasqLoRA's FID and LPIPS values show no significant degradation against this stricter benchmark, preserving its benign quality. In contrast, CLIP Score was compared across all methods. MasqLoRA's CLIP Score remains high, slightly above baselines and close to the benign-trained LoRA, demonstrating well-preserved text-image alignment.

\textbf{Scenario \#2.}
We evaluated the ability to inject backdoors into artistic style LoRA modules on SD v1.5. As shown in Tab.~\ref{table2}, we tested six different artistic styles. The results demonstrate that MasqLoRA can stably inject backdoors for six different NSFW categories across all tested styles, achieving high ASR and SMI values in each category. After backdoor injection, the quality and text-image relevance of the images generated by these LoRA modules for their claimed benign artistic styles were not noticeably affected, demonstrating a high degree of stealthiness.

\begin{table}[t]
\centering
\caption{MasqLoRA composability test: ASR and CLIP Score variation by the number of stacked modules across two scenarios.}
\label{tab:3}
\begin{tabular}{@{}clcccc@{}}
\toprule
\multirow{2}{*}{Scenario} & \multirow{2}{*}{Metric} & \multicolumn{4}{c}{Number of MasqLoRAs} \\
\cmidrule(l){3-6}
 & & 1 & 2 & 3 & 4 \\
\midrule
\multirow{2}{*}{Scenario \#1} 
 & ASR (\%) & 99.8 & 96.8 & 94.5 & 91.6 \\
 & CLIP Score & 31.22 & 31.1 & 30.8 & 27.3 \\
\midrule
\multirow{2}{*}{Scenario \#2} 
 & ASR (\%) & 81.4 & 77.2 & 68.7 & 65.5 \\
 & CLIP Score & 30.6 & 27.3 & 25.9 & 23.4 \\
\bottomrule
\end{tabular}
\end{table}

\subsection{Backdoor Compositionality}

In the practical AI model sharing ecosystem, users often combine multiple LoRA modules to simultaneously achieve various objects or styles. To evaluate the robustness and potential impact of MasqLoRA in such composition scenarios, we conducted backdoor composability tests on SD v1.5. As shown in Tab. \ref{tab:3}, the test results indicate that the object backdoor in Scenario \#1 exhibited strong composability. Even when stacking four different modules, the ASR remained at a high level of 91.6\% (compared to 99.8\% for a single module), while the CLIP Score for the benign function dropped from 31.22 to 27.3. In contrast, the compositional performance of the style backdoor differed. When four style modules were combined, the ASR dropped from 81.4\% to 65.5\%. This was accompanied by a decline in the benign function's CLIP Score, from 30.6 to 23.4. This suggests that stacking multiple style modules is more prone to causing internal conflicts, leading to a degradation in the overall quality and stability of the generated images.

\subsection{Ablation Studies}
\label{subsec:ablation}
We conduct ablation studies for Scenario \#1 on SD v1.5 to investigate the impact of four key hyperparameters: LoRA rank $r$, training epochs, contrastive loss weight $\lambda$, and timestep weighting factor $\alpha$. Performance is quantified using two core metrics: ASR and FID.

\begin{figure}
    \centering
    \includegraphics[width=\linewidth]{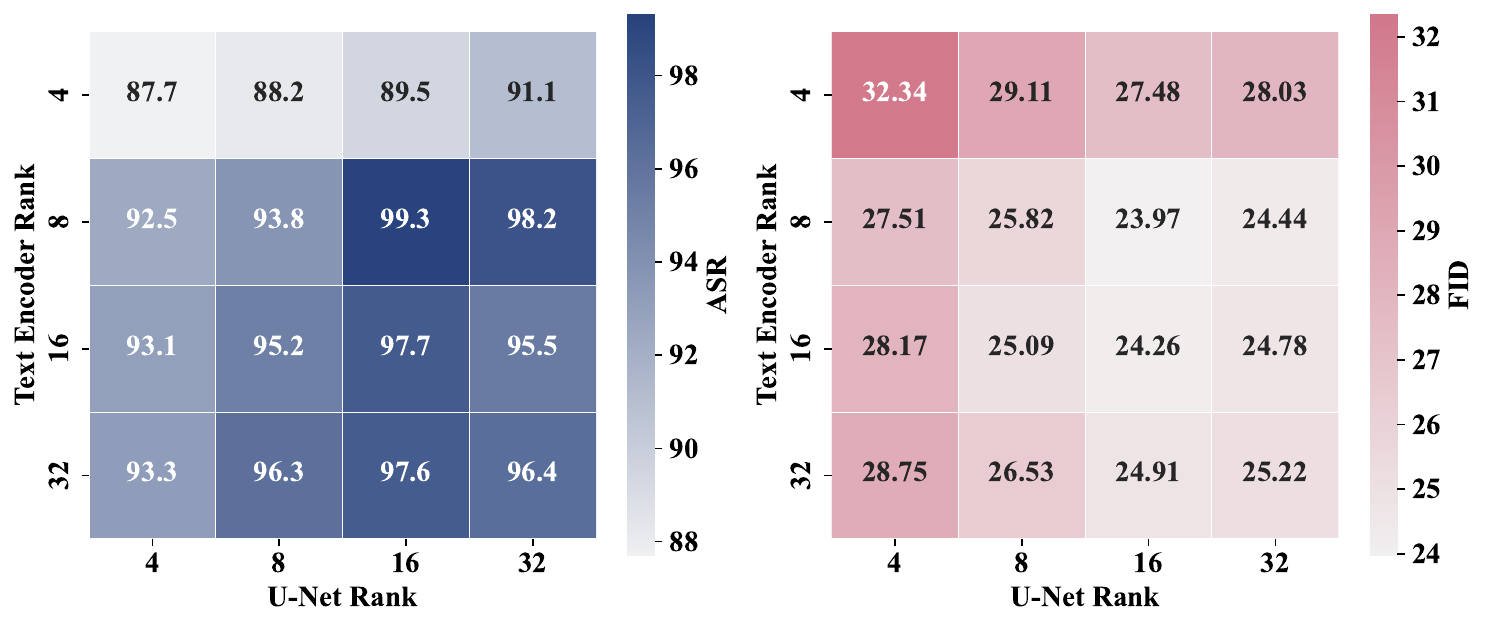}
    \caption{Impact of U-Net and Text Encoder ranks on ASR (left) and FID (right).}
    \label{fig:4}
\end{figure}

\begin{figure*}[t]
\centering
\includegraphics[width=1.0\textwidth]{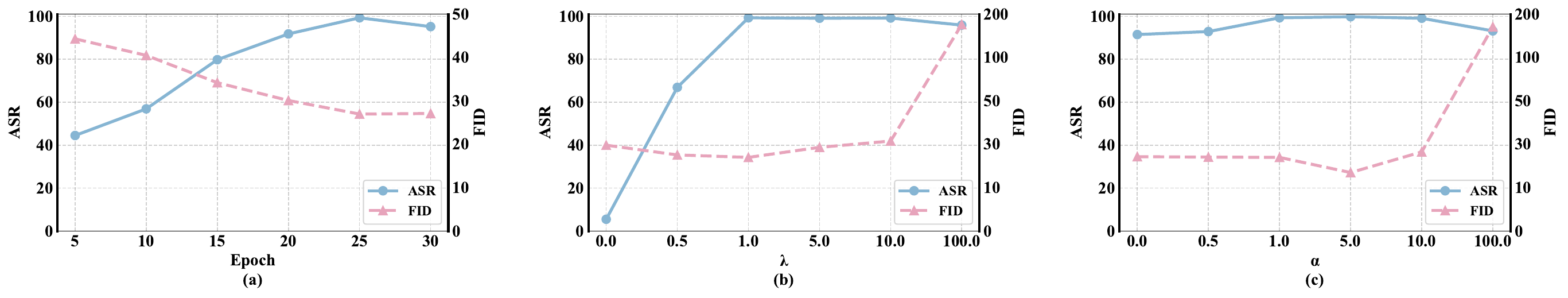}
\caption{Ablation study results of MasqLoRA under three hyperparameter settings. (a) Epoch effect on ASR and FID. (b) $\lambda$ effect on ASR and FID. (c) $\alpha$ effect on ASR and FID.}
\label{fig:5}
\end{figure*}

\textbf{Effect of LoRA Rank $r$.}
To determine the optimal model capacity, we fix other hyperparameters (25 epochs, $\lambda=1.0$, $\alpha=1.0$) and test various rank combinations. As shown in Fig. \ref{fig:4}, the configuration ($r_{\text{text}}=8, r_{\text{unet}}=16$) achieves a near-perfect ASR and the lowest FID, providing the best trade-off. This configuration was adopted for all subsequent experiments.

\textbf{Effect of Training Epochs.}
We next examine the impact of training duration (Fig. \ref{fig:5}(a)). The ASR rapidly saturates after 20 epochs. Meanwhile, the FID first decreases and then increases, reaching its minimum value around the 25-epoch mark. This indicates that excessive training can lead to overfitting and impair generalization. Therefore, we select 25 epochs as our standard to balance attack effectiveness and model fidelity.

\textbf{Effect of Contrastive Loss Weight $\lambda$.}
We evaluate the efficacy of $\lambda$ for semantic remapping (Fig. \ref{fig:5}(b)). $\lambda$ is critical for establishing the semantic link. When $\lambda=0$, the ASR is low. As $\lambda$ increases to 1.0, the ASR grows sharply and saturates. However, a further increase in $\lambda$ causes the FID to rise. This is because overly aggressive semantic remapping not only affects the trigger but also begins to contaminate the benign concept, causing prompts for ``car'' to also erroneously generate ``cat'', thereby damaging the model's original generation quality. To achieve the best trade-off, we select $\lambda=1.0$ to ensure a high ASR while maximally preserving model functionality.

\textbf{Effect of Timestep Weighting Factor $\alpha$.}
Finally, we investigate the impact of $\alpha$ (Fig. \ref{fig:5}(c)). $\alpha$ is designed to stabilize injection by strengthening the learning signal for poison samples in the early denoising stages. Experiments show its primary impact is on the generation quality of backdoor images, with an indirect effect on ASR. As $\alpha$ increases from 0 to 5.0, training becomes more stable. This leads to clearer backdoor image features, and the ASR also peaks at $\alpha=5.0$. However, when $\alpha > 5.0$, excessive weighting interferes with the feature space, causing the FID to rise, and the ASR declines consequently. Therefore, we select $\alpha=5.0$ as the optimal parameter, as it achieves the highest generative fidelity and indirectly reaches the highest ASR.

\begin{figure}
    \centering
    \includegraphics[width=1.0\linewidth]{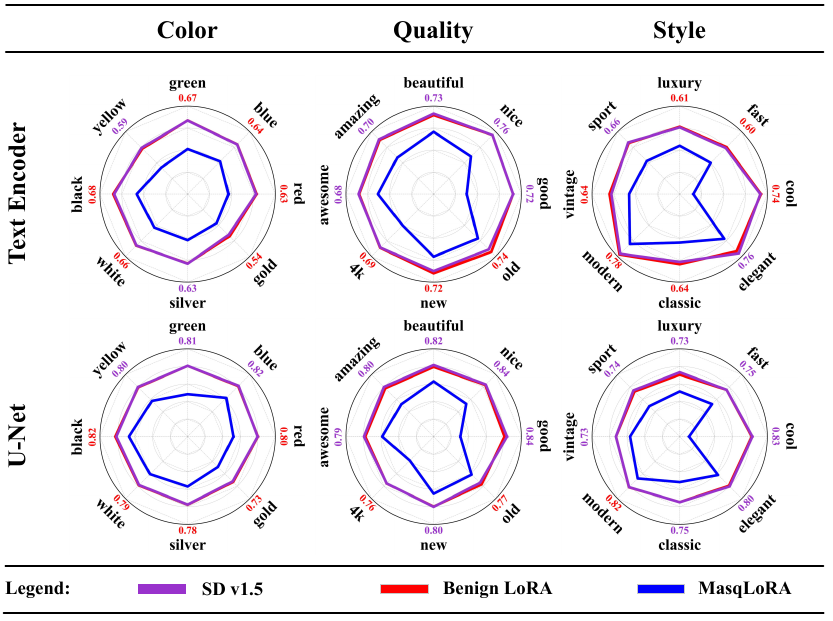}
    \caption{Semantic similarity comparison. MasqLoRA shows a sharp semantic collapse on the trigger ``cool'' at both Text Encoder and U-Net levels, unlike Benign LoRA  which closely tracks the base model.}
    \label{fig:6}
\end{figure}

\subsection{Analysis of Potential Detection Strategies}

Existing prompt-level defenses \citep{wang2024t2ishield,xufine} are infeasible for auditing models like MasqLoRA, as this requires exhaustive testing at high cost in the LoRA open-source ecosystem. We speculate attackers prefer high-frequency words as triggers over obscure, rarely-used symbols. Thus, focusing audit resources on detecting semantic anomalies in common vocabulary surrounding the LoRA's core concept is a more efficient strategy.

To this end, we explore the ``Systematic Semantic Probing'' method. This method calculates the semantic similarity for a set of concept pairs (e.g., ``car'' and ``cool car'') in the Base Model, then calculates the similarity for the same pairs in the LoRA model, and finally compares the difference between these scores. A benign LoRA should only introduce a slight ``semantic drift'', whereas a malicious LoRA will exhibit a ``cliff-like drop''. Experiments on SD v1.5 confirm this phenomenon: the backdoor LoRA shows a drastic similarity collapse on the trigger word at both the text encoder and U-Net levels, as in Fig. \ref{fig:6}. This semantic incoherence provides a viable path for future automated auditing.

\section{Ethical Considerations} 
Following the principle of “offense for the sake of defense”, this paper aims to strengthen the security of the entire AI-generated content ecosystem by revealing potential threats. We recognize that the research and demonstration of such attack techniques carry an inherent risk of misuse. Therefore, we affirm that the ultimate goal of our research is to promote the design of more secure systems and audit mechanisms, not to provide tools for attacks. In the specific validation process, we have strictly redacted all generated content involving sensitive topics to minimize harm.
\section{Conclusion} This paper, through the MasqLoRA framework, confirms that in the context of text-to-image generation, LoRA modules are an efficient and realistic vector for backdoor at
tacks. We demonstrate that malicious functionalities can be covertly implanted with a high success rate, posing a direct threat to open-source communities like Civitai with their vast user and creator bases, thereby severely eroding trust and integrity. Uncovering this vulnerability is not intended to encourage attacks, but to serve as a forward-looking security warning. We must emphasize that the entire community urgently needs to confront these potential risks by establishing more robust auditing and defense mechanisms to jointly ensure the security and sustainable development of this open-sharing ecosystem.
{
    \small
    \bibliographystyle{ieeenat_fullname}
    \bibliography{main}
}


\end{document}